\documentclass[journal]{IEEEtran}

\ifCLASSINFOpdf
  \usepackage[pdftex]{graphicx}
 
\else

\fi

\usepackage{amsmath}

\usepackage{listings}
\usepackage{tcolorbox}
\usepackage{xcolor}
\usepackage{colortbl}
\usepackage{booktabs}
\usepackage{siunitx}
\usepackage{wrapfig}
\usepackage{flafter}
\usepackage{pifont}
\usepackage[export]{adjustbox}
\usepackage{listings}
\usepackage{color}
\newcommand\code[1]             {{\bf\ttfamily\small #1}}

\definecolor{0}{HTML}{1f77b4}
\definecolor{1}{HTML}{ff7f0e}
\definecolor{2}{HTML}{2ca02c}
\definecolor{3}{HTML}{d62728}
\definecolor{4}{HTML}{9467bd}
\definecolor{5}{HTML}{8c564b}
\definecolor{6}{HTML}{e377c2}
\definecolor{7}{HTML}{7f7f7f}
\definecolor{8}{HTML}{bcbd22}
\definecolor{9}{HTML}{17becf}

\ifCLASSOPTIONcompsoc
 \usepackage[caption=false,font=normalsize,labelfont=sf,textfont=sf]{subfig}
\else
 \usepackage[caption=false,font=footnotesize]{subfig}
\fi

\usepackage{url}

\begin{document}

\title{Machine Learning for Sensor Transducer Conversion Routines}

\author{Thomas~Newton,
  James~T.~Meech,~\IEEEmembership{Student~Member,~IEEE}~%
  and~Phillip~Stanley~Marbell,~\IEEEmembership{Senior~Member,~IEEE}
  \thanks{Thomas~Newton, James~T.~Meech and Phillip~Stanley~Marbell are with the Department
    of Engineering, University of Cambridge, Cambridge, UK CB3 0FF e-mail: phillip.stanley-marbell@eng.cam.ac.uk.}
  }

\markboth{}%
{}


\maketitle

\begin{abstract}
  Sensors with digital outputs require software conversion routines to transform the unitless analogue-to-digital converter samples to physical quantities with correct units. 
These conversion routines are computationally complex given the limited computational resources of low-power embedded systems. 
This article presents a set of machine learning methods to learn new,
less-complex conversion routines that do not sacrifice accuracy for the BME680 environmental sensor.
We present a Pareto analysis of the tradeoff between accuracy and computational overhead for
the models and models that reduce the computational overhead of the existing industry-standard conversion routines for temperature, pressure, and
humidity by \SI{62}{\%}, \SI{71}{\%}, and \SI{18}{\%} respectively. The corresponding RMS errors are
\SI{0.0114}{^\circ C}, \SI{0.0280}{KPa}, and \SI{0.0337}{\%}. These results show that 
machine learning methods for learning conversion routines can produce conversion routines with reduced computational overhead which 
maintain good accuracy. 
\end{abstract}


\IEEEpeerreviewmaketitle

\section{Introduction}
\IEEEPARstart{T}{here} is a constant drive to make sensors more power-efficient.
Users require embedded sensor systems such as wearable fitness trackers and battery-powered smart thermostats to be small with long battery life.
In resource-constrained sensor applications, engineers optimise code for a small
memory footprint and low computational overhead~\cite{david2020tensorflow}.

This article describes new methods to optimise the computational overhead of the
conversion routines for the BME680, a digital temperature, pressure, humidity, and
indoor air quality sensor~\cite{bme680manual}.
Digital sensors require conversion routines to have meaningful outputs. Embedded sensor systems require low-power
microcontrollers with sufficient random-access memory (RAM), flash storage, and
computational performance to store and run the conversion routines to use sensors: 
Optimising the conversion routines will help meet these requirements under tighter memory size and power consumption constraints.

\begin{figure}[!t]
   \includegraphics[width=0.5\textwidth]{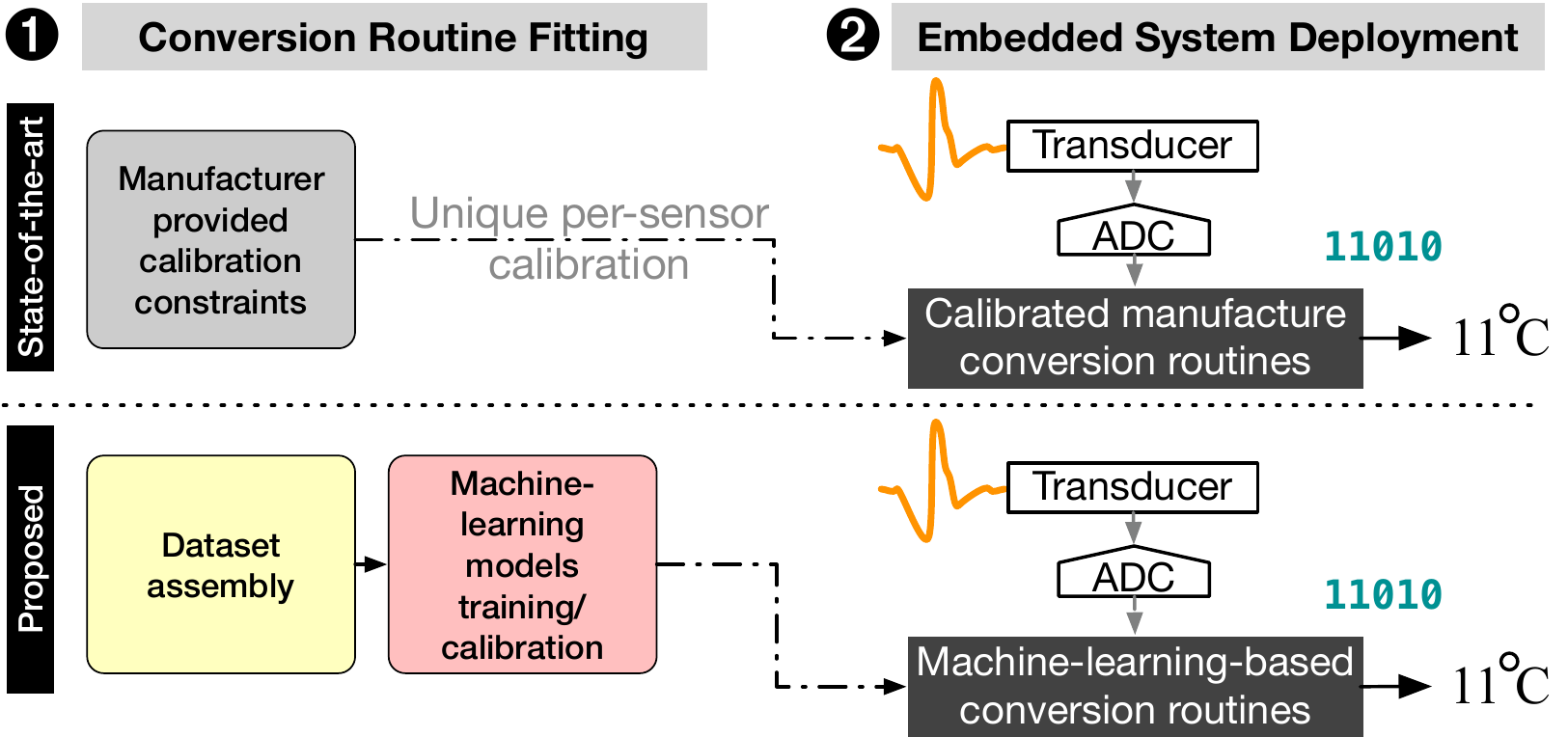}%
   \caption{Block diagrams for the original and new learned
        conversion routines.}
		\label{fig:original_block_diagram}
\end{figure}

\subsection{Contributions}
\noindent
This article presents the following three contributions: 
\begin{enumerate}
    \item A new approach and its implementation of alternative, machine-learned conversion routines for temperature,
          pressure, and humidity. 
		  We demonstrate the methods using a state-of-the-art sensor, the Bosch BME680 but the insights and methods are applicable to other sensors.
    \item Quantitative evaluation of the computational overhead, accuracy, RAM, and flash storage usage of these conversion routines.
    \item A Pareto analysis of accuracy against computational overhead for the conversion routines.
\end{enumerate}

\subsection{Why do Sensors Need Conversion Routines?}
\noindent
Typical sensors with digital outputs have a transducer which is sensitive to the
signal they are designed to measure (the measurand), e.g., temperature, and produce a voltage which is
converted to a digital signal by an analogue-to-digital converter (ADC). The ADC
outputs a single unsigned integer with a fixed number of bits (the measurement).
Obtaining a reading with meaningful units requires
a transformation to map these unsigned integers into the correct real number range.
If the transducer is non-linear then the conversion
must also invert the sensor transducer transfer function after the ADC has digitised its analogue output signal~\cite{stretchSensor}.

\subsection{Manufacturer-Provided Conversion Routines}
\noindent
The top half of Figure~\ref{fig:original_block_diagram} shows the BME680
conversions provided by Bosch Sensortec~\cite{bme680manual}. 
An external microcontroller computes the conversion routines in real time each time the microcontroller samples the sensor.
For the BME680, the manufacturer provides calibration constants
specific to each sensor. These are parameters of the conversion
routine, which tune the conversion on a per-sensor basis to reduce the
impact of any manufacturing variability in sensor properties.

\subsection{Learning New Conversions from Data}
\noindent
We can use machine learning to learn alternative conversion
routines from data. The bottom half of Figure~\ref{fig:original_block_diagram} shows a block diagram of the process.
 Using the original manufacturer-provided conversion routines and a set of
calibration constants, our method generates training datasets, to train the models.
Our method saves the trained weights so that at inference time the models are a direct replacement for the manufacturer-provided conversion routines in Figure~\ref{fig:original_block_diagram}. The calibration
constants used to generate the training data are specific to one physical
sensor, therefore the trained weights will only be accurate for that sensor.
\section{Training Data}\label{sec:training_data}
\noindent
We produced synthetic mesh datasets and sequence datasets for training and testing the models.
The synthetic datasets take the original conversion routines, provided by the
manufacturer~\cite{bme680manual}, as ground truth.
We generated mesh datasets by producing a fine mesh of equally-spaced points
across the unconverted domain. We applied the original conversion routines
to produce labels for all of these data points.

To produce datasets which are continuous sequences, we used Gaussian processes,
adjusting the characteristics using a kernel function.
We used a five-halves Matern kernel function and
set the length scale to $20$ times the virtual sampling period. The virtual
sampling period is the constant interval between evaluating the
Gaussian process and has no link to any real sampling period. To ensure that
the generated sequence fills the required range, we apply a linear
transformation after generating the function. This makes the
amplitude parameter of the kernel function irrelevant. For multi-dimensional
inputs, we evaluated a single dimensional sample function for each input.
We drew sample functions independently from the same single dimensional Gaussian process.
We applied the original conversions to provide labels for the dataset.

An optional refinement, the inverse refinement, ensures datasets
precisely match the sensor operating range. Instead of generating data
in the unconverted domain, we generate data in the converted domain, where it
is easy to enforce the limits of the operating range.
We compute the unconverted values by inverting the original conversion
routines.

\section{Evaluation Method}
\noindent
We measured the accuracy of the models using a sequence of synthetic data which was $1000$ points
long. We averaged each result over ten similar datasets.
For models that included an element of randomness such as
neural networks, we averaged the results over five different random seeds. 
\subsection{Accuracy Measure}
\noindent
Let $y_\mathrm{Predict}$ be the value predicted by the machine learning model, $y_\mathrm{True}$ be the ground truth value from the original conversion routine, and $y_\mathrm{True}^\mathrm{(Max)}$ and $y_\mathrm{True}^\mathrm{(Min)}$ be the maximum and minimum values of the operating range of the sensor for the physical quantity of interest. 
We used a normalised RMS error to quantify accuracy, where:
\small
\begin{equation}\label{eq:normalised_error}
    E_\mathrm{RMS} = \sqrt{\sum_i \left( (y_\mathrm{Predict}^{(i)}-y_\mathrm{True}^{(i)}) \times\frac{100}{y_\mathrm{True}^\mathrm{(Max)}-y_\mathrm{True}^\mathrm{(Min)} }\right)^2} .
\end{equation}
\normalsize
This allows comparison between the measurements from different sensors
(e.g., temperature -40 to 85\,$^\circ$C, pressure \SI{30}{kPa} to \SI{110}{kPa}, and humidity 0 to \SI{100}{\%}~\cite{bme680manual}).

\subsection{Computational Overhead and Memory Usage Measures}
\noindent
We first implemented each model in Python making use of the TensorFlow library~\cite{tensorflow}. 
We then exported the trained weights and created \lstinline{C}-language implementations of the inference functions based on these weights.
We measured overhead as the number of RISC-V instructions needed to
convert a single ADC sample into a physically-meaningful real-valued number with units. 
We used the Sunflower embedded system emulator~\cite{sunflower} to measure the number
of dynamic instructions executed at run time by a binary compiled from a given \lstinline{C}-language implementation:
We used a feature in Sunflower to precisely and deterministically measure the dynamic instructions
executed for a given region of source code. Computational overhead translates almost linearly to energy consumption so we use computational overhead as the cost metric~\cite{sunflower}. We measured the flash and RAM usage of the implementations of the models generated by each machine learning method by
taking the size of the \code{.text} section as the flash storage requirement. We summed the
\code{.bss} and \code{.data} section sizes to determine the RAM requirement for global
variables and manually counted the memory required for local variables.
\section{Tested Models}
\noindent
We evaluated both function approximation methods and time series methods.
\subsection{Function Approximation Methods}\label{sec:function_approximation_methods}
\noindent
We trained most models with datasets generated by the mesh method described in
Section~\ref{sec:training_data} using the inverse refinement and 20 
discretisation levels each for temperature, pressure, and humidity. 
We used other datasets where explicitly mentioned.
The methods that we trained fall into five broad categories: 

\subsubsection{Linear Regression}\label{sec:linear_regression}
\noindent
Simple linear-in-the-parameters regression using linear feature vectors.
We used a simple non-Bayesian linear regression.

\subsubsection{Quadratic Regression}\label{sec:quadratic_regression}
\noindent
A linear-in-the-parameters regression using quadratic feature vectors.

\subsubsection{Linear Interpolation Lookup Table}\label{sec:lookup_table}
\noindent
A lookup table that linearly interpolates between three nearby points.
We used bespoke datasets using the mesh method described in
Section~\ref{sec:training_data} without the inverse refinement. We did this to 
ensure a regular, square grid in the unconverted (input) domain to simplify 
the interpolation. We tested discretisation levels of 3, 10, and 20.

\subsubsection{Gaussian Process Regression}\label{sec:gaussian_processes}
\noindent
Gaussian Process regression as described by Rasmussen~\cite{gaussianProcessses}.
We used a Gaussian likelihood function and Gaussian kernel to solve the
regression analytically. The training datasets used discretisation levels of two 
and three. We optimised the hyper-parameters in the kernel function to maximise the
log-likelihood of generating the data.

\subsubsection{Neural Networks}\label{sec:neural_networks}
\noindent
Simple feed-forward fully-connected neural networks. Every neuron uses the same activation function. 
A preliminary evaluation showed that the exponential, Gaussian kernel, sigmoid, RELU, hard sigmoid, tanh, and soft sign activation functions had dynamic instruction overheads of 109, 131, 122, 19, 33, 124, and 11 RISC-V instructions respectively. We therefore chose to use the RELU function because it is
commonly used and has low computational overhead (19 dynamic instructions). We trained models to
minimise the RMS error (Equation~\ref{eq:normalised_error}). To aid training, we rescaled the inputs and outputs to the range $[0, 1]$.

\subsection{Sequence-Based Models}
\noindent
Sequence-based models may be able to provide lower computational overhead
by taking advantage of the fact that the sensors measure physical systems and
we have prior knowledge of what a sequence of sensor data might look like.

We trained all the sequence-based models to minimise the RMS error.
We generated $5000$ point training datasets using the
sequence method from Section~\ref{sec:training_data}. Training required
approximately $20,000$ epochs. By comparison, the feed-forward
neural networks required fewer than $10,000$ epochs.
We trained four categories of sequence-based models:

\subsubsection{Auto-Regressive Moving Average (ARMA)}\label{sec:arma}
\noindent
ARMA models are usually used in systems with many wide sense stationarity
assumptions and where the inputs are white noise. With these
assumptions the optimal parameters can be
computed analytically but in this case, with no simplifying assumptions, we
trained them iteratively using TensorFlow.

\subsubsection{Gated Recurrent Unit (GRU)}\label{sec:gru}
\noindent
A single GRU cell with a history vector of length one,
providing a scalar output as required. There are two different
versions of the GRU: We used the version proposed by Chung et al.~\cite{gru}, but both
usually perform similarly~\cite{gru}.
Long short term memory (LSTM) neurons, used by Oldfrey et al.~\cite{stretchSensor} are
more complicated than the GRU and have higher computational overhead.

\subsubsection{Half GRU}\label{sec:half_gru}
\noindent
A GRU cell without the output gate. The output is returned directly after applying the activation function.
This reduces computational overhead.

\subsubsection{Simple RNN}\label{sec:simple_rnn}
\noindent
The simple RNN is the least-complex of all, with no gating. The non-linear activation function
allows it to model more complex data than the ARMA models.
\section{Experimental Evaluation}
\noindent
For each of the models increasing the number of input dimensions increases the computational overhead, the humidity routines are slightly more complex than the those for pressure because they clamp the output between 0 and 100\,\%.
The sequence-based methods introduce additional computational overhead without providing a corresponding decrease in error. 
More complicated sensor conversion routines using sequence-based methods could produce a reduction in the error that justifies the increased overhead.
Machine learning models are most uncertain (or over confident) where they have the least training data~\cite{foong2019inbetween}.
We trained the function approximation methods using a mesh of points that uniformly span the domain.
The model never had to generalise outside of the training data and therefore overfitting is not a problem. 
For sequence-based methods the performance on real data and synthetic training and test data was similar indicating that overfitting was not a problem.

\noindent
\textbf{Temperature conversion routines:} Figure~\ref{fig:pareto_plots}(a) shows the Pareto plot of error and computational
overhead for the temperature conversion routine. Quadratic regression (\textcolor{2}{\ding{68}})
Pareto dominates the original conversion (its normalised RMS error
was of the order $10^{-12}$, rounded to $0$) and its computational overhead is
\SI{47}{\%} lower than the original. Linear regression (\textcolor{1}{\ding{67}})
achieves a \SI{62}{\%} reduction in overhead compared to the original.
There is a small increase in normalised RMS error, from approximately $10^{-12}$ to
$0.00909$, which corresponds to an RMS error of \SI{0.0114}{^\circ C}.
Figure~\ref{fig:resource_usage}(a) shows that the linear and quadratic
regression also have lower RAM and flash storage usage than any other
method including the original.

\noindent
\textbf{Pressure conversion routines:} Figure~\ref{fig:pareto_plots}(b) shows that the pressure
Pareto frontier includes the original (\textcolor{0}{\ding{66}}), the linear interpolation lookup
table with 400 entries (\textcolor{2}{\ding{78}}), the quadratic regression (\textcolor{2}{\ding{68}}), and the linear regression (\textcolor{1}{\ding{67}}). 
The original conversion has zero error by definition since we used it as ground
truth. The linear interpolation lookup table with 400 entries (\textcolor{2}{\ding{78}}) has the next smallest
error of $0.0223$ with an overhead reduction of \SI{49}{\%}.
Although it results in an increase in normalised error of $0.0350$ (i.e., \SI{28.0}{Pa}), quadratic regression (\textcolor{2}{\ding{68}})
offers a larger reduction in computational overhead of \SI{71}{\%} compared to
the original. Linear regression (\textcolor{1}{\ding{67}}) offers
a greater reduction in overhead of \SI{85}{\%} but with
increased error of up to $0.998$ which is \SI{798}{Pa}.
Figure~\ref{fig:resource_usage}(b) shows that the linear and quadratic regression
have lower resource usage than the original but the linear interpolation lookup table with 400 entries (\textcolor{2}{\ding{78}}) has a high RAM usage of
\SI{1652}{bytes} compared to \SI{60}{bytes} for the original conversion routine. This is because
it stores a lookup table consisting of $400$ floating-point numbers in RAM.
Storing the lookup table in flash storage would be possible but it would
increase the overhead due to the need to move the data between flash storage
and memory.

\def \pareto_chart_height{1.45in}
\begin{figure}[!t]
    \subfloat[Temperature.]{\adjincludegraphics[trim=0 0 {0.615\width} 0, clip, height=\pareto_chart_height]{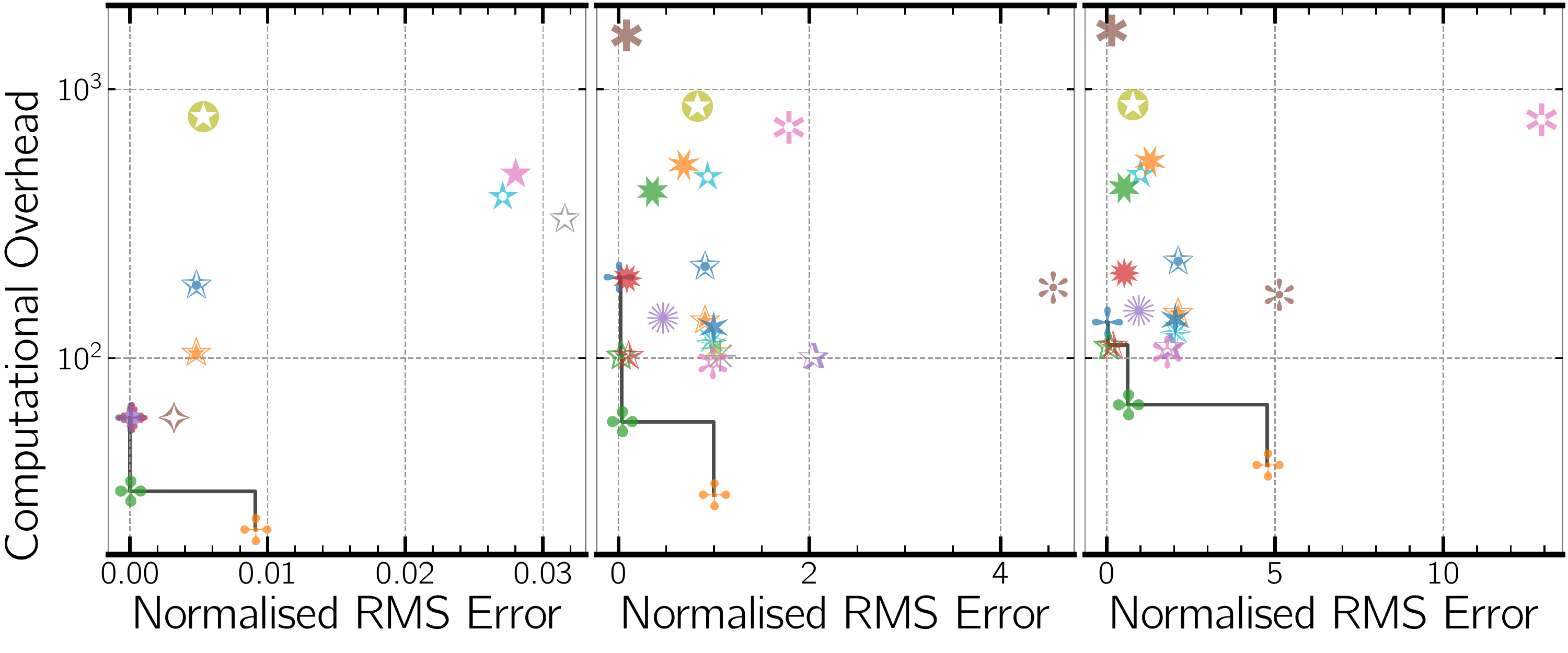}%
        \label{fig:temperature_pareto}}
    \subfloat[Pressure.]{\adjincludegraphics[trim={0.385\width} 0 {0.31\width} 0, clip, height=\pareto_chart_height]{Figures/pareto_chart.pdf}%
        \label{fig:pressure_pareto}}
    \subfloat[Humidity.]{\adjincludegraphics[trim={0.69\width} 0 0 0, clip, height=\pareto_chart_height]{Figures/pareto_chart.pdf}%
        \label{fig:humidity_pareto}}
\vspace{-0.05in}
	\begin{tcolorbox}[colframe=white,colback=white]
		\resizebox{\textwidth}{!}{%
		\begin{tabular}{ll}
		\textcolor{0}{\ding{66}:~Original} & \textcolor{1}{\ding{67}:~Linear Regression} \\
		\textcolor{2}{\ding{68}:~Quadratic Regression} & \textcolor{3}{\ding{69}:~Linear Interpolation LUT 20} \\
		\textcolor{4}{\ding{70}:~Linear Interpolation LUT 10} & \textcolor{5}{\ding{71}:~Linear Interpolation LUT 3}\\
		\textcolor{6}{\ding{72}:~Gaussian Process 3} & \textcolor{7}{\ding{73}:~Gaussian Process 2} \\
		\textcolor{8}{\ding{74}:~Neural Network 3-3-1} & \textcolor{9}{\ding{75}:~Neural Network 3-1} \\
		\textcolor{0}{\ding{76}:~Neural Network 1-1} &  \textcolor{1}{\ding{77}:~Neural Network 1} \\
		\textcolor{2}{\ding{78}:~Linear Interpolation LUT 400} & \textcolor{3}{\ding{79}:~Linear Interpolation LUT 100} \\
		\textcolor{4}{\ding{80}:~Linear Interpolation LUT 9} & \textcolor{5}{\ding{81}:~Gaussian Process 9} \\
		\textcolor{6}{\ding{82}:~Gaussian Process 4} & \textcolor{7}{\ding{83}:~AR 1 MA 1} \\
		\textcolor{8}{\ding{84}:~AR 2 MA 1} & \textcolor{9}{\ding{85}:~AR 3 MA 1} \\
		\textcolor{0}{\ding{86}:~AR 3 MA 2} & \textcolor{1}{\ding{87}:~GRU 1 tanh sigmoid} \\
		\textcolor{2}{\ding{88}:~GRU 1 RELU sigmoid} & \textcolor{3}{\ding{89}:~GRU 1 RELU softsign} \\
		\textcolor{4}{\ding{90}:~Half GRU 1 RELU softsign} & \textcolor{5}{\ding{91}:~Simple RNN 1 tanh} \\
		\textcolor{6}{\ding{92}:~Simple RNN 1 RELU} &  \\
		\end{tabular}}
    \end{tcolorbox}
\vspace{-0.25in}
    \caption{
	\label{fig:pareto}
	Pareto plots for each conversion routine where smaller overhead and normalised RMS error is desirable. Pareto
    frontiers marked in black. See the caption of Figure~\ref{fig:resource_usage} for an explanation of the omitted results.}
	\vspace{-0.1in}
    \label{fig:pareto_plots}
\end{figure}

\noindent
\textbf{Humidity conversion routines:} Figure~\ref{fig:pareto_plots}(c) shows the Pareto plot for the humidity conversion.
The Pareto frontier includes the same four methods as for the pressure: original (\textcolor{0}{\ding{66}}),
the linear interpolation lookup table with 400 entries (\textcolor{2}{\ding{78}}), quadratic (\textcolor{2}{\ding{68}}) and linear (\textcolor{1}{\ding{67}})
regression. As before the original conversion (\textcolor{0}{\ding{66}}) has an error of $0$. The linear interpolation lookup table with 400 entries (\textcolor{2}{\ding{78}}) has normalised error of $0.0344$ and an overhead reduction of \SI{18}{\%}. 
Quadratic regression (\textcolor{2}{\ding{68}}) provides a larger
reduction in overhead of \SI{51}{\%} with a normalised error of $0.626$. This error is
much larger than that for the quadratic regression for the pressure conversion.
Linear regression (\textcolor{1}{\ding{67}}) provides the best computational overhead with a
reduction of \SI{71}{\%} compared to the original but the normalised error is high
at $4.77$. Figure~\ref{fig:resource_usage}(c) shows that the RAM usage of the linear interpolation lookup table with 400 entries (\textcolor{2}{\ding{78}}) is high, but the other methods on the Pareto frontier have lower resource usage than the original.
If we exclude the linear interpolation lookup table with 400 entries (\textcolor{2}{\ding{78}}), a smaller lookup table
replaces it on the Pareto frontier. Most methods had larger error on the humidity conversion than on the pressure
conversion. The 400 entry linear interpolation lookup table (\textcolor{2}{\ding{78}}) was an exception.

\def\resource_chart_height{2.03in}
\begin{figure}[!t]
    \subfloat[Temperature.]{\adjincludegraphics[trim=0 0 {0.65\width} 0, clip, height=\resource_chart_height]{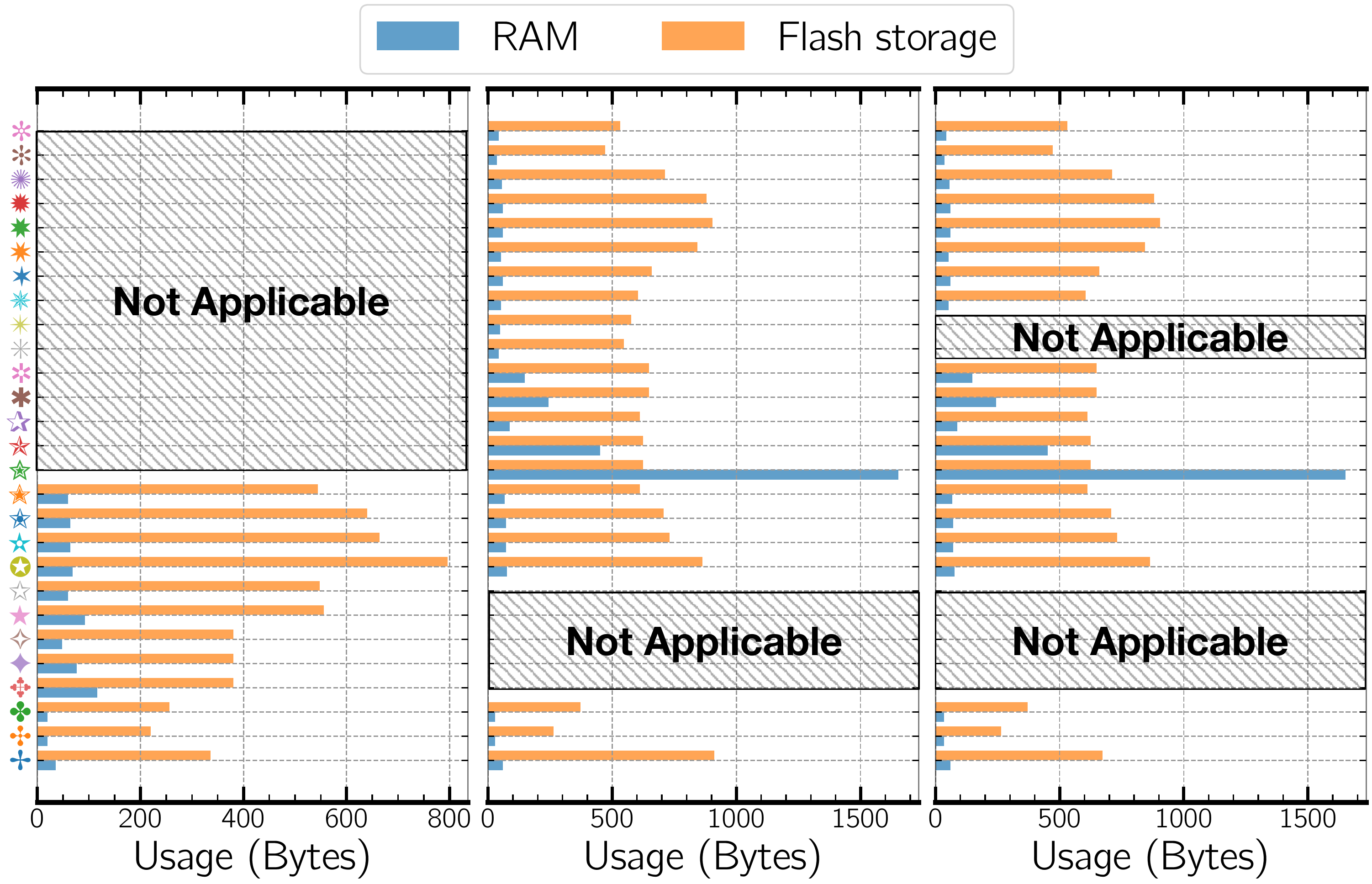}%
        \label{fig:temperature_resource}}
    \subfloat[Pressure.]{\adjincludegraphics[trim={0.35\width} 0 {0.33\width} 0, clip, height=\resource_chart_height]{Figures/resource_usage.pdf}%
        \label{fig:pressure_resource}}
    \subfloat[Humidity.]{\adjincludegraphics[trim={0.67\width} 0 0 0, clip, height=\resource_chart_height]{Figures/resource_usage.pdf}%
        \label{fig:humidity_resource}}
    \caption{RAM and flash storage usage of each model. The caption of Figure~\ref{fig:pareto_plots} shows the key for the y-axis.
			 The hatched regions correspond to models which are not appropriate for the conversion routine being modelled or where we had prior knowledge that the quadratic regression would Pareto dominate all the sequence based methods. 
	\vspace{-0.1in}}
    \label{fig:resource_usage}
\end{figure}

\section{Related Work}
\noindent
As far as we are aware there is no prior work on using machine learning to find
conversion routines with reduced computational overhead and memory
requirements for an existing, commercially-available sensor. Two recent review
papers discuss the use of machine learning for signal processing on the same
physical hardware as the sensor but this requires bespoke hardware with
integrated sensing and computing capability~\cite{zhou2020near, ballard2021machine}. Our method
of learning new sensor conversion routines from data can run on existing
hardware without modification. Oldfrey et al. present a machine learning approach
that linearises the output of cheap, highly-non-linear stretch
sensors~\cite{stretchSensor} and Zhang et al. reconstruct a full
spectrum from reduced data~\cite{spectrometer}. Neither of these methods are
directly applicable to the learning of sensor conversion routines for existing
commercially-available sensors from data.
\section{Conclusion}
\noindent
Lookup tables with linear interpolation, quadratic and linear regression
are all superior to the original conversion routines in terms of computational overhead. The Pareto frontiers for the conversions we tested were
entirely occupied by function approximation methods: Sequence-based methods ranging from ARMA to RNNs, LSTMs, and GRUs, were not on the Pareto frontier. The function approximation methods avoid the overheads of sequence
models including initialising internal state, the possibility that error will
be worse on some unusual sequences, and the possibility that models are
unstable.

For temperature, linear regression provides a reduction in
computational overhead of \SI{62}{\%} with an error of \SI{0.0114}{C}. For pressure,
quadratic regression provides a \SI{71}{\%} reduction in
computational overhead and an error of \SI{0.0280}{KPa}. For humidity, a linear
interpolation lookup table with 400 data points performs well with a \SI{18}{\%}
reduction in computational overhead and an error of \SI{0.0337}{\%}.

Computational overhead reductions translate almost linearly to energy savings and are therefore
useful in low-power systems using the BME680 sensor. There is no obvious reason why these methods will not 
work with similar sensors and their conversion routines.

\section*{Acknowledgements}
\noindent
This research is supported by an Alan Turing Institute award TU/B/000096 under EPSRC grant EP/N510129/1, by EPSRC grant EP/V047507/1, and by the UKRI Materials Made Smarter Research Centre (EPSRC grant EP/V061798/1).
We thank Vasileios Tsoutsouras and Orestis Kaparounakis for assistance with the figures.

\ifCLASSOPTIONcaptionsoff
  \newpage
\fi

\vspace{-0.15in}
\bibliography{ms}
\bibliographystyle{IEEEtran}

\end{document}